\documentclass[journal,twoside,web]{ieeecolor}
\usepackage{generic}
\usepackage{cite}
\usepackage{amsmath,amssymb,amsfonts}
\usepackage{algorithmic}
\usepackage{graphicx}
\usepackage{algorithm,algorithmic}
\usepackage{hyperref}
\hypersetup{hidelinks=true}
\usepackage{textcomp}

\usepackage{tabularx}
\usepackage{multirow}
\usepackage{pifont}
\usepackage{array}
\usepackage{booktabs}
\usepackage{balance}
\usepackage{booktabs}
\usepackage{mathrsfs}
\usepackage{bm}
\usepackage[dvipsnames]{xcolor}

\renewcommand\arraystretch{1.2}
\definecolor{newcolor}{rgb}{.8,.349,.1}

\newcolumntype{M}[1]{>{\centering\arraybackslash}m{#1}}

\newcommand{\etal}{\textit{et al}. }

\newcommand{\eg}{\textit{e}.\textit{g}., }

\def\BibTeX{{\rm B\kern-.05em{\sc i\kern-.025em b}\kern-.08em
    T\kern-.1667em\lower.7ex\hbox{E}\kern-.125emX}}
\begin{document}
\title{SSiT: Saliency-guided Self-supervised Image Transformer for Diabetic Retinopathy Grading}
\author{Yijin Huang, Junyan Lyu, Pujin Cheng, Roger Tam, Xiaoying Tang
\thanks{This study was supported by the Shenzhen Basic Research Program (JCYJ20190809120205578); the National Natural Science Foundation of China (62071210); the Shenzhen Science and Technology Program (RCYX20210609103056042); the Shenzhen Basic Research Program (JCYJ20200925153847004); the Shenzhen Science and Technology Innovation Committee (KCXFZ2020122117340001); the National Key Research and Development Program of China (2023YFC2415400). (Corresponding authors: Dr. Xiaoying Tang; Dr. Roger Tam).}
\thanks{Yijin Huang is with the Department of Electronic and Electrical Engineering, Southern University of Science and Technology, Shenzhen 518055, China, and also with School of Biomedical Engineering, The University of British Columbia, Vancouver, BC V6T 1Z3, Canada (e-mail: yijinh@student.ubc.ca).}
\thanks{Junyan Lyu is with the Department of Electronic and Electrical Engineering, Southern University of Science and Technology, Shenzhen 518055, China, and also with Queensland Brain Institute, The University of Queensland, St Lucia QLD 4072, Australia (e-mail: junyan.lyu@uq.edu.au).}
\thanks{Pujin Cheng is with the Department of Electronic and Electrical Engineering, Southern University of Science and Technology, Shenzhen 518055, China (e-mail: 12032946@mail.sustech.edu.cn).}
\thanks{Roger Tam is with School of Biomedical Engineering, The University of British Columbia, Vancouver, BC V6T 1Z3, Canada (e-mail: roger.tam@ubc.ca).}
\thanks{Xiaoying Tang is with the Department of Electronic and Electrical Engineering, Southern University of Science and Technology, Shenzhen 518055, China, and also with Jiaxing Research Institute, Southern University of Science and Technology, Jiaxing 314001, China (e-mail: tangxy@sustech.edu.cn).}}

\maketitle

\begin{abstract}
Self-supervised Learning (SSL) has been widely applied to learn image representations through exploiting unlabeled images. However, it has not been fully explored in the medical image analysis field. In this work, Saliency-guided Self-Supervised image Transformer (SSiT) is proposed for Diabetic Retinopathy (DR) grading from fundus images. We novelly introduce saliency maps into SSL, with a goal of guiding self-supervised pre-training with domain-specific prior knowledge. Specifically, two saliency-guided learning tasks are employed in SSiT: (1) Saliency-guided contrastive learning is conducted based on the momentum contrast, wherein fundus images' saliency maps are utilized to remove trivial patches from the input sequences of the momentum-updated key encoder. Thus, the key encoder is constrained to provide target representations focusing on salient regions, guiding the query encoder to capture salient features. (2) The query encoder is trained to predict the saliency segmentation, encouraging the preservation of fine-grained information in the learned representations. To assess our proposed method, four publicly-accessible fundus image datasets are adopted. One dataset is employed for pre-training, while the three others are used to evaluate the pre-trained models’ performance on downstream DR grading. The proposed SSiT significantly outperforms other representative state-of-the-art SSL methods on all downstream datasets and under various evaluation settings. For example, SSiT achieves a Kappa score of 81.88\% on the DDR dataset under fine-tuning evaluation, outperforming all other ViT-based SSL methods by at least 9.48\%.
\end{abstract}


\section{Introduction}
Diabetic Retinopathy (DR) is one of the microvascular complications of diabetes and the leading cause of blindness in the working-age population of developed countries \cite{CHEUNG2010124}. Delayed treatment may induce irreversible vision impairments and malfunctions. DR biomarkers can be identified from fundus images, including hemorrhages, exudates, microaneurysms, and retinal neovascularization. However, due to the nearly imperceptible early pathological signs and the rapid increase in the number of patients with diabetes, DR screening is time-consuming and labor-intensive, even for well-trained clinicians. Therefore, automated DR detection methods are desired to reduce the number of untreated patients and the burden on clinicians, especially in regions with limited medical resources.

During the past decade, deep learning has achieved great success in the DR detection realm \cite{li2021applications}. Specifically, various Convolutional Neural Networks (CNNs) have been proposed for automated DR grading \cite{huang2023identifying}. Recently, Vision Transformers (ViTs) \cite{dosovitskiy2020image} have further boosted the performance of deep learning, exhibiting prominence on a variety of medical image modalities such as computed tomography \cite{tang2022self, hatamizadeh2022unetr} and X-ray \cite{park2022multi}. ViTs are generally more data-hungry than common CNNs \cite{dosovitskiy2020image,matsoukas2021time}, and thus large-scale datasets with high-quality annotations are required to train ViTs well. However, annotating medical images is extremely time-intensive and error-prone, exerting a heavy burden on clinical experts.

Self-supervised Learning (SSL) has been explored to learn representations from images with no annotations. Contrastive learning \cite{chen2020simple, chen2021empirical} is one of the promising SSL paradigms, wherein differently augmented views are created from the same images, and then representations are learned by maximizing the similarity between features from those different views. SSL has successfully established its effectiveness in computer vision, but the medical image analysis realm has not fully benefited from such advances yet, mainly because of the giant domain gap between natural images and medical images. In natural images, salient objects generally occupy a large portion and their characteristics are discriminative (\eg shape and color). In contrast, medical images of the same modality have similar anatomy and intensity profiles, being inadequate for disease discrimination. In such context, contrastive SSL approaches that highly rely on global informatics are prone to learn feature representations that are useful for contrastive tasks but not sufficiently discriminative for downstream medical tasks. Furthermore, medical images (\eg fundus images) may have various diagnostic features (\eg lesions) dispersed throughout the entire image. Therefore, local fine-grained information is highly crucial for medical image based disease discrimination.

To improve the performance of SSL for fundus images, more attention needs to be paid to salient regions and the learned representations are desired to preserve fine-grained information. In this work, \textbf{S}aliency-guided \textbf{S}elf-supervised \textbf{i}mage \textbf{T}ransformer (SSiT) is proposed by introducing saliency maps into SSL. A saliency map can clearly characterize the foreground of a fundus image, including the optic disc/cup, vessels, as well as lesions. A pixel-level saliency detection method requiring no training \cite{montabone2010human} is employed to obtain the saliency maps of the fundus images from a pre-training dataset without annotations. Then, as shown in Fig. \ref{fig:framework}, a contrastive learning framework is designed based on the momentum contrast \cite{chen2021empirical}, wherein two encoders, a query encoder and a key encoder, are utilized to generate representations from differently augmented views of the same input image. The key encoder is a momentum-based moving average of the learnable query encoder. To encourage the query encoder to learn representations from salient regions, trivial patches are removed from the input sequences of the key encoder as guided by the corresponding saliency maps. In this way, the key encoder guides the training of the query encoder by providing target representations that focus on salient regions. Moreover, to learn fine-grained semantics in fundus images, another pre-training objective is introduced by predicting saliency segmentation using the query encoder. Consequently, our proposed SSiT can not only explicitly guide the pre-training model to learn saliency information of the fundus images, but also allow the learned representations to preserve local fine-grained information. To validate SSiT, the EyePACS dataset\footnote{https://www.kaggle.com/c/diabetic-retinopathy-detection} without annotations is adopted to pre-train ViTs, and three publicly-accessible datasets (namely DDR \cite{ddr}, Messidor-2 \cite{messidor} and APTOS2019\footnote{https://www.kaggle.com/c/aptos2019-blindness-detection}) to evaluate the quality of the learned representations. Both quantitative and qualitative experiments demonstrate the superiority of SSiT over state-of-the-art (SOTA) self-supervised methods. Furthermore, the self-attention maps from the self-supervised ViTs in SSiT are found to display semantic information of DR-related diagnostic regions.

Our main contributions are summarized as follows:
\begin{itemize}
	\item A novel SSL framework named SSiT is proposed for DR grading from fundus images. Two learning objectives are adopted in SSiT, a saliency-guided contrastive loss and a saliency map segmentation loss. The saliency-guided contrastive loss encourages the encoder to aggregate features from salient regions, and the saliency map segmentation loss pushes the encoder to preserve fine-grained details in the learned representations. The source code is available at \href{https://github.com/YijinHuang/SSiT}{https://github.com/YijinHuang/SSiT}.
	\item Our self-supervised ViTs explicitly learn semantic information of DR-related diagnostic regions (\eg lesions and vessels), which is infeasible for other SSL methods. The rich saliency information can be clearly identified by visualizing the self-attention maps from the self-supervised ViTs in SSiT.
	\item To the best of our knowledge, this is the first work to demonstrate that saliency maps can significantly improve SSL's pre-training performance for medical images.
	\item Extensive experiments are conducted on four fundus image datasets, demonstrating that SSiT consistently outperforms representative SOTA self-supervised approaches on all datasets and under all evaluation settings.
\end{itemize}

This paper is organized as follows: Section \ref{section2} analyzes related works on DR grading, self-supervised learning, and saliency-guided vision transformer. Section \ref{section3} describes details of our proposed SSiT. In Section \ref{section4}, extensive experiments are conducted to evaluate the performance of SSiT on seven downstream datasets and under various evaluation settings. Finally, discussion and conclusion are respectively presented in Sections \ref{section5} and \ref{section6}.

\section{Related Works}
\label{section2}
\subsection{Deep Learning for DR Grading}
According to the unified and standard International Clinical Diabetic Retinopathy Scale, DR can be classified into five grades: 0 (normal), 1 (mild DR), 2 (moderate DR), 3 (severe DR), and 4 (proliferative DR) \cite{lin2020sustech}. DR-related biomarkers can be identified from fundus images, including hemorrhages, exudates, microaneurysms and retinal neovascularization. In recent years, supervised deep learning approaches have been applied to DR grading utilizing fundus images. With the powerful capability of high-level feature learning, CNNs are generally adopted as the feature extraction module in those deep learning based DR grading methods. CABNet \cite{cabnet} equips its model with two extra modules, one for exploring region-wise features and one for exploiting global attention feature maps. Huang \etal \cite{huang2023identifying} systematically analyze the impact of various deep learning components and boosts the DR grading performance based on experimental observations. Zhou \etal \cite{zhou2019collaborative} propose a collaborative learning method to jointly train disease grading and lesion segmentation by semi-supervised learning. Sun \etal \cite{sun2021lesion} propose a lesion-aware transformer for simultaneous DR grading and lesion discovery using an encoder-decoder structure with an attention mechanism.

Recently, ViTs are gaining increasing interest in computer vision and medical image analysis, yielding exceptional performance on a variety of image recognition tasks. Wu \etal \cite{wu2021vision} analyze different ViT architectures for DR grading, showing that ViTs are competitive alternatives to CNNs. Matsoukas \etal \cite{matsoukas2021time} assess the performance of ViTs for various medical image analysis tasks including DR grading, demonstrating that ViTs outperform CNNs for large-scale datasets. Yu \etal \cite{yu2021mil} improve the performance of ViTs for fundus image classification by exploiting feature representations extracted by individual image patches. Despite this, the application of ViTs to medical image analysis is still relatively limited because of the lack of well-annotated data and thus ViTs have not been fully explored yet. In this work, we aim to develop an SSL framework for ViTs to enhance the DR grading performance by leveraging unlabeled fundus images.

\subsection{Self-supervised Learning in Natural Images}
SSL has achieved overwhelming success in computer vision. As a common practice in SSL, a pretext task is designed by leveraging supervisory signals from the input image itself, which is then employed to train a neural network of interest to learn image representations. Instance discrimination \cite{chen2020simple, he2020momentum} is one of the most promising SSL paradigms in computer vision. Its learning objective is to distinguish each image from others by maximizing the similarity of representations from differently augmented views of the same image. MoCo \cite{he2020momentum} learns representations by measuring the similarity of embedded features from a trained encoder and a dynamic representation dictionary from a momentum encoder. SimCLR \cite{chen2020simple} trains encoders by measuring the similarity of images in a large batch. MoCo-v3 \cite{chen2021empirical} improves the performance of MoCo for self-supervised ViTs. DINO \cite{caron2021emerging} proposes a self-supervised ViT to learn representations using self-distillation, which trains a learnable student ViT to predict the embeddings of a momentum teacher ViT. 

These methods achieve superior performance on natural image classification tasks. However, several works \cite{haghighi2022dira, li2021mst} suggest that instance discrimination methods rely highly on global features, and thus have limited power in capturing fine-grained information. To address this problem, DenceCL \cite{wang2021dense} proposes a pairwise contrastive similarity loss at the pixel level between two views of the same input image. EsViT \cite{li2021efficient} presents a new pre-training task of region matching to capture fine-grained region dependencies. Furthermore, masked image modeling \cite{he2022masked, xie2022simmim} has gained great research attention as an SSL paradigm for ViTs. MAE \cite{he2022masked} presents a masked autoencoder for representation learning, which masks random patches from the input image and trains an encoder to reconstruct the masked patches. However, such methods are generally less competitive in direct discriminative representation learning tasks, as assessed by linear evaluation and $k$-NN classification \cite{he2022masked, wang2022semantic}.

In this work, to train a network that not only captures global and discriminative features but also preserves local and fine-grained features for fundus images, both image-level and pixel-level discriminative learning are integrated into SSiT.

\subsection{Self-supervised Learning in Medical Images}
It is super expensive to annotate large-scale medical image datasets. Therefore, many efforts have been made in SSL for medical images \cite{cai2022uni4eye}. In the ophthalmic image analysis realm, Holmberg \etal \cite{holmberg2020self} train a self-supervised model by predicting OCT-based retinal thickness from fundus images. Li \etal \cite{li2020self} propose an SSL framework to exploit multi-modal data for retinal disease diagnosis. Cai \etal \cite{cai2022uni4eye} propose a Unified Patch Embedding module in ViT for jointly processing 2D and 3D retinal images. Chen \etal \cite{chen2023unsupervised} pre-train models to focus on tiny structures by jointly training a clustering module to generate segmentation and an embedding module to embed semantically consistent pixels. Our previously proposed lesion-based contrastive learning pipeline \cite{huang2021lesion} takes lesion patches as the input to encourage the network to learn more discriminative features for DR grading. For other types of medical imaging modalities, PCRL \cite{zhou2021preservational} enhances representations learned from a contrastive loss by reconstructing diverse contexts. DiRA \cite{haghighi2022dira} combines discriminative, restorative and adversarial learning in a unified manner. Medical image-report joint pre-training \cite{wang2022multi, cheng2023prior} is another direction to learn robust and general medical vision representations. However, these methods require extra medical report data.

SSiT distinguishes itself from previous works by utilizing saliency to guide the training of SSL. In SSiT, two saliency-driven objectives are proposed to encourage a ViT model to learn representations encoded with DR-related features.

\subsection{Saliency-guided Vision Transformer}
Several studies have leveraged the informative content embedded in saliency maps to enhance the training or feature extraction processes of ViTs. Zhu \etal \cite{zhu2021saliency} introduce a saliency-guided transformer, TranSLA, for image quality assessment. TranSLA combines saliency prediction with the Transformer architecture, directing the model's attention to regions of interest. Similarly, Lu \etal \cite{lu2023saliency} propose a saliency-guided ViT for few-shot keypoint detection, wherein a saliency map acts as a soft mask, guiding the self-attention mechanisms to focus on foreground elements. 

However, these saliency-guided ViTs are mainly designed for supervised settings, while SSiT focuses on the self-supervised learning setting. Moreover, existing saliency-guided ViTs are mainly designed for natural images, with limited validation of the reliability of saliency maps in the context of medical images. To leverage the saliency map for guiding ViTs' training under a self-supervised setting using fundus images, SSiT adopts a distinctive approach. Unlike other saliency-guided transformers that employ saliency map as additional features to enhance the input information, SSiT utilizes it as a hint to generate self-supervised training targets for saliency-aware representation learning.

\begin{figure*}[tbp]
	\centering
	\includegraphics[width=0.95\linewidth]{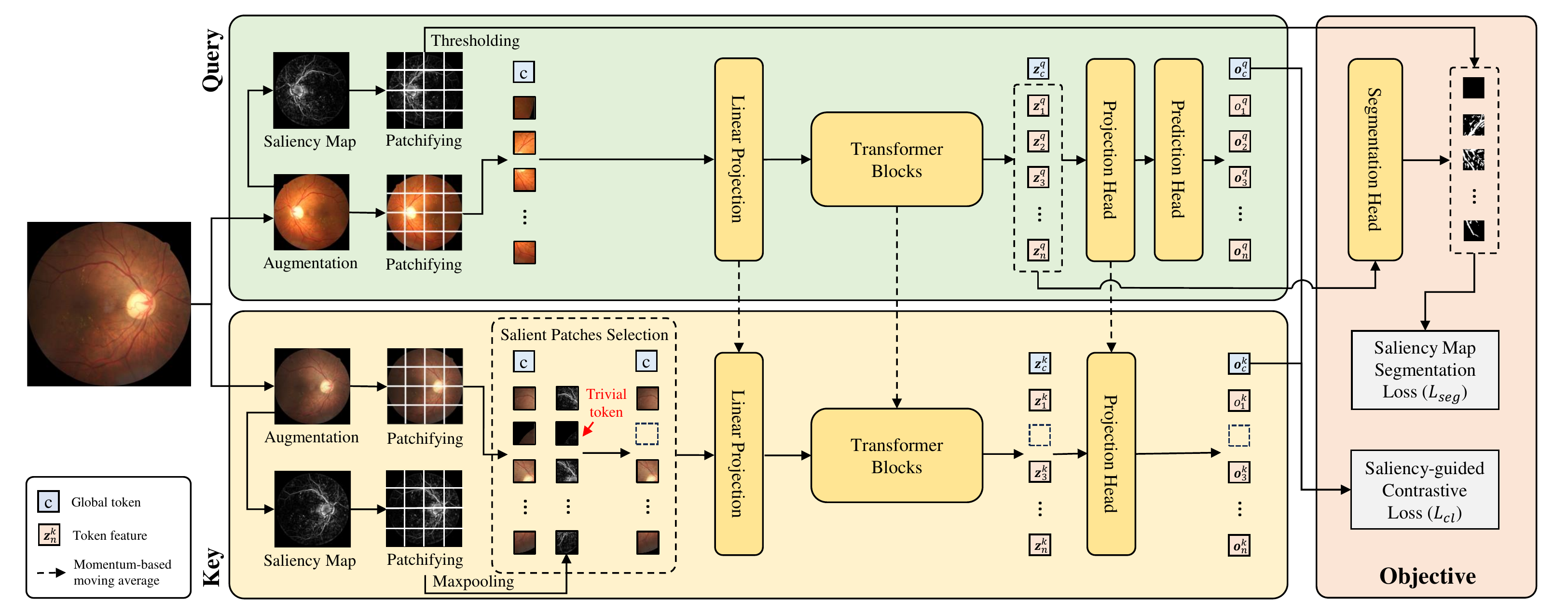}
	\caption{The proposed SSiT framework. There are two learning objectives, a saliency-guided contrastive loss and a saliency segmentation loss. First, different augmentations of the same input image are fed into two encoders, a query encoder and a key encoder. The two encoders have the same architecture, and the parameters of the key encoder are updated by the moving average of the learnable query encoder. Salient patches selection is performed for the input sequence of the key encoder to remove trivial patches, thus constraining the key encoder to provide target representations focusing on salient regions. The query encoder is optimized by maximizing the similarity between the output representation and the target. In addition, the query encoder is also trained to directly predict saliency segmentation to learn fine-grained information from fundus images.}
	\label{fig:framework}
\end{figure*}

\section{Methodology}
\label{section3}
\subsection{Vision Transformer}
In this section, we briefly introduce the mechanism of ViT \cite{vaswani2017attention, dosovitskiy2020image}. ViT can be treated as a feature extractor that encodes representations, which is a competitive alternative to CNN. A standard ViT consists of a linear projection layer and multiple transformer blocks. Given an image $x \in \mathbb{R}^{H \times W \times C}$ with spatial resolution $(H, W)$ and number of channels $C$, we first patchify it to non-overlapping 2D patches of size $P \times P$, where $P$ is defined as the patch size of the ViT of interest. Then, the 2D patches are flattened and mapped into a latent embedding space of dimension $D$, via the linear projection layer. This results in patch embeddings $\{\bm{z}_p^i \in \mathbb{R}^{D} \,|\, i = 1,...,N\}$, where $N = HW / P^2$ denotes the sequence length. A learnable embedding $\bm{z}_{class}$ is prepended to the sequence of patch embeddings. $\bm{z}_{class}$ serves as the global representation token to aggregate information from the entire sequence. Furthermore, to retain the spatial position information, learnable position embeddings $\textbf{E}_{pos}$ are added to the patch embeddings. The resulting sequence, which is the input to the transformer blocks, is presented as follows
\begin{equation}
	\bm{z}_{0} = [z_{class};\, z_{p}^{1};\, ...;\, z_{p}^{N}] + \textbf{E}_{pos}.
\end{equation}
Unlike convolution-based feature extractors, ViTs model long-range information through self-attention \cite{vaswani2017attention}. The encoder of a ViT comprises alternating transformer blocks each consisting of a multi-headed self-attention (MSA) layer and a multi-layer perceptron (MLP) block. Given a ViT of $L$ layers, the input sequence is processed as follows
\begin{align}
	&\bm{z}_{\ell}^{\prime} = \text{MSA}(\text{LN}(\bm{z}_{\ell-1})) + \bm{z}_{\ell-1}, \\
	&\bm{z}_{\ell} = \text{MLP}(\text{LN}(\bm{z}_{\ell}^{\prime})) + \bm{z}_{\ell}^{\prime},
\end{align}
where $\ell$ indexes the layers and $\text{LN}(\cdot)$ denotes the layer normalization operation \cite{ba2016layer}. In an MSA layer, each patch embedding updates itself by aggregating other patch embeddings according to a computed self-attention map. The final output consists of a global representation $\bm{z}_{L, class}$ and patch representations $\{\bm{z}_{L, i} \,|\, i = 1,...,N\}$. The global representation $\bm{z}_{L, class}$ is generally considered as the representation of the image of interest, which can be further attached to other task-specific layers (\eg a fully connected layer for classification).

\subsection{Saliency-guided Contrastive Learning}
\subsubsection{Vanilla Contrastive Learning}
\label{vcl}
As a common practice in contrastive learning, random compositions of data augmentation operations (\eg cropping, color jittering and flipping) are applied to an input image to generate two different views. Following \cite{grill2020bootstrap, chen2021empirical}, the two different views are then encoded by two encoders, a query encoder $f_q$ and a key encoder $f_k$, which are respectively parameterized by $\theta_q$ and $\theta_k$. The query encoder consists of a ViT, a projection head and a prediction head, while the key encoder has the same architecture except for the prediction head. The key encoder is a momentum encoder, whose parameters are updated by a moving average of the parameters of the query encoder. Specifically,
\begin{equation}
	\theta_{k} = \alpha \theta_{k} + (1 - \alpha) \theta_{q},
\end{equation}
where $\alpha$ is defined as the momentum coefficient. Given an input image, the two encoders output features $o_q$ and $o_k$. Contrastive learning is formulated to maximize the similarity between features from the two views of the same input image and minimize that from different images. In this work, the InfoNCE loss \cite{oord2018representation} is adopted as the objective function
\begin{equation}
	\mathcal{L}_{cl} = - \log \frac{\text{exp}(o_{q} \cdot o_{k}^{+} / \tau)}{\text{exp}(o_{q} \cdot o_{k}^{+} / \tau) + \sum\limits_{o_{k}^{-}} \text{exp}(o_{q} \cdot o_{k}^{-} / \tau)},
\end{equation}
wherein positive sample $o_{k}^{+}$ denotes features from the same source image as $o_{q}$, while negative sample set $\{o_{k}^{-}\}$ denotes features from other images in the same training batch. $\tau$ is a temperature parameter \cite{wu2018unsupervised} controlling the sharpness of the contrastive loss. When pre-training is finished, only the ViT in the query encoder $f_q$ is kept and the features $\bm{z}_{L, class}$ before the projection head are treated as the image representation.

\subsubsection{Guidance of Saliency Map}
\label{gsm}
Although vanilla contrastive learning has shown effectiveness in the natural image realm, it has limited power in capturing salient information from fundus images, which is subtle, diverse and dispersed over the entire image. To encourage the self-supervised encoder to mainly focus on salient regions, saliency maps are introduced into contrastive learning. As shown in Fig. \ref{fig:framework}, a saliency map highlights the conspicuous regions in an image of interest, in which the pixel value represents the degree of saliency. A static saliency detection method \cite{montabone2010human} is adopted in this work, which computes saliency based on center-surround differences of the image, and thus it can be applied to medical images with intensity differences between regions of interest and background, such as fundus images, OCT and MRI. This method is employed to generate the saliency maps of all fundus images in a pre-training dataset.

In our aforementioned contrastive learning scheme, a momentum encoder is involved, which plays a vital role in providing target features for a learnable query encoder. To effectively and efficiently teach the learnable query encoder, trivial patches from the input sequence of the momentum encoder are removed according to the saliency map, thereby providing target representations that focus on salient regions. As shown in Fig. \ref{fig:framework}, the saliency map is first patchified similar to that for the input image, and take the maximum saliency value in each patch as the patch-wise saliency score. Then, $m$\% patches with the lowest saliency scores are removed from the input sequence of the momentum encoder, wherein the masking ratio $m$ is a hyperparameter controlling the number of patches to remove. In this way, the momentum encoder is constrained to provide target representations corresponding to the salient patches. By maximizing the feature similarity between positive samples from the momentum encoder, the query encoder learns to aggregate features from salient patches. Note that patch exclusion is only performed for the momentum encoder during pre-training, and thus it would not affect the transfer capability of the query encoder that uses the entire image as the input in the downstream DR grading task.

\subsection{Saliency Map Segmentation}
Contrastive learning generally formulates the pretext task as an image-level discriminative prediction problem making use of global image representations. Therefore, it has very limited power in preserving fine-grained details which are nevertheless essential for fundus image based DR diagnosis. In this context, an additional pixel-level saliency map segmentation task is proposed to train the model to capture subtle and local information. The saliency maps are first binarized via thresholding to generate the ground truth saliency masks $\bm{y}$ for segmentation. To reconstruct the segmentation map at the full resolution as the input image, a lightweight decoder is appended to the final transformer block of the query encoder, taking patch representations $\{\bm{z}_{L, i} \in \mathbb{R}^{D} \,|\, i = 1,...,N\}$ as the input. The decoder consists of a linear layer followed by sigmoid activation, mapping each $D$-dimension patch representation to a $P^2$-dimension feature vector. All $P^2$-dimension feature vectors are reshaped back to the original patch size $P \times P$. We then recover the spatial order of the patch segmentation and concatenate all patch segmentation to form the final segmentation prediction $\bm{\hat{y}}$ with the full resolution $H \times W$. The saliency map segmentation loss is computed as the cross entropy loss function
\begin{equation}
	\mathcal{L}_{seg}(\bm{y}, \bm{\hat{y}}) = \frac{1}{\Omega(\bm{y})} \sum_{i=1}^{\Omega(\bm{y})} -\bm{y_{i}} \log (\bm{\hat{y}_{i}}),
\end{equation}
where $\Omega(\cdot)$ denotes the total number of pixels. To accurately segment the saliency map, the encoder is encouraged to learn the shape, color and texture of the salient regions, and thus benefits the identification of abnormal regions when transferring to the downstream DR grading task. The decoder is removed after pre-training.

\subsection{Joint Training}
Finally, the total objective of SSiT is formulated as below
\begin{equation}
	\mathcal{L} = \lambda_{cl} \ast \mathcal{L}_{cl} + \lambda_{seg} \ast \mathcal{L}_{seg},
\end{equation}
where $\lambda_{cl}$ and $\lambda_{seg}$ are hyper-parameters balancing the two objectives. To minimize the objective $\mathcal{L}_{cl}$, the query encoder is required to capture the saliency of the input image and project the output representation into the latent space encoded by the saliency-guided momentum encoder. Meanwhile, $\mathcal{L}_{seg}$ encourages the query encoder to learn salient characteristics and encode fine-grained information into representations by training the model to predict pixel-level saliency segmentation.

Collectively, SSiT can effectively leverage saliency maps to guide the query encoder to learn representations encoded with saliency information and fine-grained details.

\begin{table*}[t]
	\centering
	\caption{Comparison results with state-of-the-art self-supervised learning methods on the three evaluation datasets. The best ones are bolded while the second best ones are underlined. [$\kappa$ (\%)]}
	\label{table:compare}
	\renewcommand{\arraystretch}{1.2}
	\begin{tabular}{l@{\hspace{0.32cm}}l@{\hspace{0.32cm}}c@{\hspace{0.34cm}}c@{\hspace{0.34cm}}cc@{\hspace{0.34cm}}c@{\hspace{0.34cm}}cc@{\hspace{0.34cm}}c@{\hspace{0.34cm}}c}
		\toprule
		\multirow{2}{*}{Method}          & \multirow{2}{*}{Arch.}  & \multicolumn{3}{c}{DDR}                                                                          & \multicolumn{3}{c}{Messidor-2}                                                                   & \multicolumn{3}{c}{APTOS2019}                                                                    \\ \cmidrule(lr){3-5} \cmidrule(lr){6-8} \cmidrule(lr){9-11}
		&& Fine-tuning                       & Linear                             & $k$-NN                    & Fine-tuning                       & Linear                             & $k$-NN                    & Fine-tuning                       & Linear                             & $k$-NN                    \\ \hline
		\it{CNNs} &&&&&&&&&& \\
		\textcolor{gray}{Random init.}   & \textcolor{gray}{VG16} & \textcolor{gray}{26.54 $\pm$ 1.21} & \textcolor{gray}{19.80 $\pm$ 1.38}  & \textcolor{gray}{24.86} & \textcolor{gray}{33.09 $\pm$ 2.47} & \textcolor{gray}{25.46 $\pm$ 1.82} & \textcolor{gray}{20.19}  & \textcolor{gray}{65.87 $\pm$ 1.43} & \textcolor{gray}{39.84 $\pm$ 1.55}  & \textcolor{gray}{62.66} \\
		\textcolor{gray}{Random init.}   & \textcolor{gray}{RN18} & \textcolor{gray}{24.31 $\pm$ 2.10} & \textcolor{gray}{17.03 $\pm$ 3.56}  & \textcolor{gray}{21.63} & \textcolor{gray}{24.30 $\pm$ 2.78} & \textcolor{gray}{16.43 $\pm$ 3.38} & \textcolor{gray}{9.54}  & \textcolor{gray}{57.90 $\pm$ 2.17} & \textcolor{gray}{30.51 $\pm$ 2.66}  & \textcolor{gray}{56.27} \\
		\textcolor{gray}{Random init.}   & \textcolor{gray}{RN50} & \textcolor{gray}{26.76 $\pm$ 2.11} & \textcolor{gray}{21.95 $\pm$ 3.17}  & \textcolor{gray}{23.37} & \textcolor{gray}{24.06 $\pm$ 3.00} & \textcolor{gray}{22.60 $\pm$ 3.95} & \textcolor{gray}{11.74}  & \textcolor{gray}{63.79 $\pm$ 1.35} & \textcolor{gray}{35.21 $\pm$ 2.40}  & \textcolor{gray}{64.17} \\
		LD \cite{chen2023unsupervised}       & VG16               & 70.79 $\pm$ 0.84                                  & 47.12 $\pm$ 2.71                                  & 32.87                       & 59.70 $\pm$ 3.47                                  & 20.60 $\pm$ 3.38                                  & 19.57                       & 88.09 $\pm$ 0.30                                  & 67.37 $\pm$ 1.09                                  & 65.32                        \\
		Reconstruction                             &RN18               & 64.09 $\pm$ 1.07                                  & 26.87 $\pm$ 2.33                                 & 24.49                       & 31.71 $\pm$ 4.07                                  & 23.38 $\pm$ 2.12                                  & 24.10                       & 84.70 $\pm$ 0.20                                  & 40.72 $\pm$ 0.89                                 & 67.52                        \\
		PCRL \cite{zhou2021preservational}                              & RN18               & 71.91 $\pm$ 0.46                                  & 54.20 $\pm$ 1.44                                  & 41.21                       & 66.63 $\pm$ 3.35                                  & 34.40 $\pm$ 1.59                                  & 19.35                       & 89.41 $\pm$ 0.39                                  & 75.47 $\pm$ 0.55                                  & 71.27                        \\
		MoCo-v2 \cite{chen2020improved}       & RN50               & 72.41 $\pm$ 0.80                                  & 60.02 $\pm$ 1.10                                  & 43.12                       & 65.67 $\pm$ 1.01                                  & 47.03 $\pm$ 1.35                                  & 33.96                       & 90.20 $\pm$ 0.33                                  & 81.82 $\pm$ 0.16                                  & 74.34                        \\
		DenseCL \cite{wang2021dense}        & RN50               & 76.96 $\pm$ 0.51                                  & 59.20 $\pm$ 0.47                                  & 45.31                       & 70.94 $\pm$ 0.92                                  & 20.15 $\pm$ 3.46                                  & 35.19                       & 91.15 $\pm$ 0.63                                  & 81.36 $\pm$ 0.14                                  & 75.58                        \\
		DiRA \cite{haghighi2022dira}                              & RN50               & \underline{78.68 $\pm$ 0.29}                     & 51.86 $\pm$ 1.01                                   & 25.99                       & \underline{74.07 $\pm$ 1.69}                                 & 24.44 $\pm$ 0.72                                  & 15.48                       & \underline{92.34 $\pm$ 0.36}                                  & 69.80 $\pm$ 0.51                                  & 59.92                        \\

		\hline
		\multicolumn{2}{l}{\it{ViTs}} &&&&&&&&& \\
		\textcolor{gray}{Random init.}   & \textcolor{gray}{ViT-S} & \textcolor{gray}{18.40 $\pm$ 7.09} & \textcolor{gray}{15.56 $\pm$ 5.60}  & \textcolor{gray}{19.42} & \textcolor{gray}{19.55 $\pm$ 2.40} & \textcolor{gray}{15.80 $\pm$ 7.14} & \textcolor{gray}{5.06}  & \textcolor{gray}{70.87 $\pm$ 0.59} & \textcolor{gray}{57.99 $\pm$ 2.19}  & \textcolor{gray}{56.48} \\
		SimCLR \cite{chen2020simple}      & ViT-S                      & 72.15 $\pm$ 0.83                                  & \underline{67.69 $\pm$ 1.68}                       & 53.53                                  & 65.47 $\pm$ 1.45                                  & 59.28 $\pm$ 2.02                       & \underline{40.87}                                  & 90.34 $\pm$ 0.29                                  & \underline{86.79 $\pm$ 0.60}                      & \underline{83.29} \\
		MoCo-v3 \cite{chen2021empirical}      & ViT-S                   & 71.41 $\pm$ 1.61                                  & 66.08 $\pm$ 1.15                                  & \underline{56.05}                       & 63.46 $\pm$ 0.74                                  & 59.82 $\pm$ 2.05                                  & 40.41                       & 90.15 $\pm$ 0.58                                  & 85.19 $\pm$ 0.53                                  & 80.68                        \\
		DINO \cite{caron2021emerging}         & ViT-S                   & 68.49 $\pm$ 0.94                                  & 66.38 $\pm$ 0.53                                  & 49.86                       & 60.98 $\pm$ 0.92                                   & \underline{60.47 $\pm$ 2.39}                                  & 40.65                       & 90.50 $\pm$ 0.32                                  & 85.07 $\pm$ 0.37                                  & 79.44                       \\
		MAE \cite{he2022masked}               & ViT-S             & 72.40 $\pm$ 1.91                                  & 59.31 $\pm$ 0.89                                  & 38.70                       & 51.35 $\pm$ 4.28                                   & 43.50 $\pm$ 1.59                                  & 21.39                       & 90.97 $\pm$ 0.35                                  & 83.24 $\pm$ 0.54                                  & 75.64                       \\
		EsViT \cite{li2021efficient}                                 & ViT-S             & 71.54 $\pm$ 0.53                                  & 59.89 $\pm$ 1.53                                  & 49.01                       & 63.74 $\pm$ 1.00                                   & 51.84 $\pm$ 1.62                                  & 38.97                       & 91.03 $\pm$ 0.34                                  & 80.42 $\pm$ 1.13                                  & 75.18                       \\
		SSiT (ours)                   & ViT-S      & \textbf{81.88 $\pm$ 0.26}                                   & \textbf{71.33 $\pm$ 0.78}                                  & \textbf{58.89}                       & \textbf{77.53 $\pm$ 0.84}                                  & \textbf{67.23 $\pm$ 0.53}                                  & \textbf{49.42}                       & \textbf{92.97 $\pm$ 0.29}                                  & \textbf{89.65 $\pm$ 0.20}                                  & \textbf{84.65}                       \\
		\bottomrule
	\end{tabular}
\end{table*}

\subsection{Implementation Details}
\label{train_detail}
\subsubsection{Vision transformer}
Following the design in \cite{touvron2021training, rw2019timm}, a ViT of moderate size, namely ViT-S (ViT Small), is adopted in this work in all experiments. ViT-S consists of 12 transformer blocks each with a 6-headed self-attention module and a 384-dimension hidden size for embedding features. The patch size is $16 \times 16$, resulting in a sequence of length 196 for a $224 \times 224$ image. There are a total of 21M trainable parameters in ViT-S.
\subsubsection{Data augmentation}
To generate different views in contrastive learning, random cropping is first applied to each input image. Specifically, for the query encoder, a cropping magnitude is randomly sampled in the range $(8\%, 80\%)$ of the original image size. For the momentum encoder, a cropping magnitude is randomly sampled in the range $(80\%, 100\%)$, ensuring that the momentum encoder provides relatively global target representations. Then, all cropped images are resized to $224 \times 224$, followed by combinations of random horizontal/vertical flipping, random rotation, color distortion and Gaussian blurring. For the image cropping strategies employed in other compared methods, we follow their default settings from the original papers. The same data augmentation strategy as SSiT is employed in all other compared methods excluding DINO \cite{caron2021emerging} for which we adopt the multi-crop augmentation strategy \cite{caron2020unsupervised, grill2020bootstrap} following its original design.
\subsubsection{Pre-training setting}
All SSL models are pre-trained from scratch. SSiT is trained using the AdamW optimizer\cite{loshchilov2017decoupled} with an initial learning rate of $1 \times 10^{-3}$ and a weight decay of 0.1. The model is trained for 300 epochs with a mini-batch size of 512 on four NVIDIA RTX 3090 GPUs. Learning rate warmup \cite{goyal2017accurate} is adopted in the first 40 epochs, and then a cosine annealing schedule during the remaining epochs. The momentum coefficient $\alpha$ also follows a cosine schedule ranging from 0.99 to 1.0. The temperature parameter $\tau$ is set to be 0.2. The masking ratio $m$\% in the momentum encoder is empirically set to be $25$\%. The objective balancing parameters $\lambda_{cl}$ and $\lambda_{seg}$ are respectively set to be 1 and 10. To ensure fair comparisons across different methods, grid searching is applied to hyper-parameters of other compared methods around the default settings from the original papers. The hyper-parameters that we search for include learning rate, weight decay and momentum coefficient where applicable.

\section{Experimental Results}
\label{section4}
\subsection{Evaluation Setting}
\label{eval_details}
To evaluate the effectiveness of the learned representations of each pre-trained model, three widely used evaluation protocols are employed for the DR grading task, with the same evaluation metric.
\subsubsection{Evaluation protocols}
Following previous works \cite{chen2021empirical, caron2021emerging, he2022masked}, three common evaluation protocols are adopted, namely fine-tuning evaluation, linear evaluation and $k$-NN classification \cite{wu2018unsupervised}, to assess the performance of each pre-trained model. For the fine-tuning evaluation, the model employed in the downstream task is initialized with pre-trained parameters of the self-supervised model of interest and then trained in a fully-supervised manner with the corresponding training set of each evaluation dataset. In contrast, in linear evaluation, the parameters from the self-supervised model are frozen and only a linear classification head is trained under a supervised downstream task setting. As for the $k$-NN classification, all image representations are extracted from each pre-trained model, and then classification is performed with a non-parametric $k$-NN classifier. Here $k$ is empirically set to be 10 in all experiments.

In general, fine-tuning evaluation assesses the transfer capability of a pre-trained model, whereas linear evaluation and $k$-NN classification directly assess the quality of the learned representations. Under all evaluation settings, the global representation $\bm{z}_{L, class}$ and the average of the patch representations $\frac{1}{N} \sum_{i=1}^{N} \bm{z}_{L, i}$ of an image of interest are concatenated and presented as the overall image representation for the downstream DR grading task. Following common practice \cite{kolesnikov2020big}, all fine-tuning and linear evaluation experiments run at a $384 \times 384$ resolution, except that the linear evaluation of MoCo-v3 \cite{chen2021empirical} runs at a $256 \times 256$ resolution because its performance deteriorates when the input resolution further increases. We conjecture that because ViT's parameters of the linear projection layer are fixed in MoCo-v3, the features cannot be properly extracted when there are sharp changes in resolution. For the fine-tuning evaluation and the linear evaluation, each experiment is repeated 5 times and report the mean $\pm$ standard deviation values. There is no randomness in the $k$-NN classification, and thus there is only one value for each experiment.

\subsubsection{Evaluation metric}
The class distribution of a DR-related fundus image dataset is usually extremely unbalanced, wherein images of each DR grade are dramatically less than normal images. Therefore, keeping in line with previous works \cite{cabnet, huang2021identifying, sun2021lesion}, we here adopt an officially-employed metric in the Kaggle DR grading competition, namely the quadratically weighted kappa score $\kappa$ \cite{cohen1968weighted}, to evaluate the downstream DR grading performance. Different from accuracy, $\kappa$ can effectively measure the multi-class classification performance on an imbalanced dataset. $\kappa$ ranges from -1 to 1, with 1 indicating complete agreement between the prediction and the groundtruth and -1 complete disagreement.

\subsection{Datasets}
\subsubsection{Pre-training dataset}
The EyePACS dataset is adopted as the pre-training dataset in this work, consisting of 88,702 fundus images with different DR grades. The images were acquired from 44,351 patients with different imaging devices and under a variety of imaging conditions, resulting in great image diversity in the dataset. Note that there is no annotation involved in the pre-training phase.

\subsubsection{Evaluation datasets}
To analyze the learned representations and the transfer learning capacity of each pre-trained model for DR grading, three evaluation datasets (DDR, Messidor-2 and APTOS2019) are employed in this work.

\textbf{DDR.} The DDR dataset consists of 13,673 fundus images. Six-category annotations (five DR grades and one ``ungradable" category) are provided for this dataset. All ungradable images are excluded following \cite{cabnet}, ending up with 6,320, 2,503 and 3,759 images for training, validation and testing.

\textbf{Messidor-2.} The Messidor-2 dataset consists of 1748 fundus images with five DR grades and eye pairing annotations. The dataset is randomly divided into three splits (60\%/10\%/30\%) for training/validation/testing, while keeping images from the same patient in the same split. The limited number of images for training in the fine-tuning evaluation and linear evaluation settings is the main challenge of the Messidor-2 dataset.

\textbf{APTOS2019.} A total of 5590 fundus images with five DR-grading annotations are provided in the APTOS2019 dataset. However, only annotations of the training set (3,662 images) are publicly-accessible, for which we randomly split into 70\%/15\%/15\% for training/validation/testing.

\subsection{Comparisons with State-of-the-art}
To demonstrate the effectiveness of SSiT, we compare it with representative SOTA SSL methods on the three evaluation datasets across different CNN and ViT architectures. As shown in Table \ref{table:compare}, poor DR grading performance is observed using models initialized randomly. Significant improvements are observed in all SSL methods compared to random initialization. Under fine-tuning evaluation, SSiT shows strong transfer capability and generalizability, consistently outperforming all other ViT-based SSL methods by at least $9.48$\% on DDR, $12.06$\% on Messidor-2 and $2.00$\% on APTOS2019. Moreover, SSiT outperforms DiRA, the previous best method with a ResNet architecture, by 3.20\% on DDR, 3.46\% on Messidor-2 and 0.63\% on APTOS2019, with a much shorter training schedule. From linear evaluation and $k$-NN classification, it is also observed that our method exceeds all other compared methods, implying that SSiT can better learn feature representations with diagnostic power for DR grading. In particular, on Messidor-2 with a training set of only 1048 samples, SSiT outperforms SOTA SSL methods by at least $12.06$\%, $6.25$\% and $8.55$\% respectively for fine-tuning, linear evaluation and $k$-NN classification, which clearly identifies the superiority of SSiT for providing robust and transferable representations in limited data regimes. In summary, experimental results suggest that SSiT is a comprehensive SSL approach that encourages the encoder to learn effective and generalizable representations, and thus can greatly enhance the downstream DR grading performance by leveraging unlabeled fundus images and utilizing the saliency information of the fundus images.

\subsection{Ablation Studies}
In this section, ablation studies are conducted to further investigate the importance of different components in SSiT. Analysis on the contribution of each objective is presented in Section \ref{role} below. If not specified, all experiments are conducted with the same hyper-parameters as described in Sec. \ref{train_detail} and Sec. \ref{eval_details}.

\begin{table}[t]
	\centering
	\caption{The fine-tuning evaluation performance of SSiT with different masking ratios. [$\kappa$ (\%)]}
	\label{table:maskratio}
	\renewcommand{\arraystretch}{1.2}
	\begin{tabular}{cccc}
		\toprule
		Masking ratio            & DDR                       & Messidor-2                  & APTOS2019                 \\ \hline
		0\%                     & 79.98 $\pm$ 1.19          & 79.48 $\pm$ 1.42   & 92.28 $\pm$ 0.38          \\
		25\%                  & \textbf{81.88 $\pm$ 0.26} & 77.53 $\pm$ 0.84            & \textbf{92.97 $\pm$ 0.29} \\
		50\%                  & 81.04 $\pm$ 0.79          & \textbf{79.97 $\pm$ 1.21}            & 92.62 $\pm$ 0.30          \\
		75\%                & 78.33 $\pm$ 1.00          & 76.09 $\pm$ 1.77            & 92.52 $\pm$ 0.40          \\
		\bottomrule
	\end{tabular}
\end{table}

\begin{table*}[t]
	\centering
	\caption{The performance of SSiT on all three evaluation datasets with different ViT architectures. [$\kappa$ (\%)]}
	\label{table:network1}
	\renewcommand{\arraystretch}{1.2}
	\begin{tabular}{lccccccccc}
		\toprule
		\multirow{2}{*}{Architecture}          & \multicolumn{3}{c}{DDR}                                                                          & \multicolumn{3}{c}{Messidor-2}                                                                   & \multicolumn{3}{c}{APTOS2019}                                                                    \\ \cmidrule(lr){2-4} \cmidrule(lr){5-7} \cmidrule(lr){8-10}
		& Fine-tuning                       & Linear                             & $k$-NN                    & Fine-tuning                       & Linear                             & $k$-NN                    & Fine-tuning                       & Linear                             & $k$-NN                    \\ \hline
		ViT-Ti          & 76.91 $\pm$ 1.17                                  & 68.69 $\pm$ 1.17                       & 55.91                                  & 74.34 $\pm$ 2.09                                  & 61.58 $\pm$ 0.63                       & 43.75                                  & 92.53 $\pm$ 0.40                                  & 87.86 $\pm$ 0.25                      & 82.19 \\
		ViT-S          & 81.88 $\pm$ 0.26                                   & 71.33 $\pm$ 0.78                                  & 58.89                       & 77.53 $\pm$ 0.84                                  & 67.23 $\pm$ 0.53                                  & 49.42                       & \textbf{92.97 $\pm$ 0.29}                                  & 89.65 $\pm$ 0.20                                  & 84.65                       \\
		ViT-B          & \textbf{83.75 $\pm$ 0.98}                                  & \textbf{73.06 $\pm$ 0.51}                                  & \textbf{60.30}                       & \textbf{80.01 $\pm$ 1.67}                                   & \textbf{69.56 $\pm$ 1.11}                                  & \textbf{54.67}                       & 92.59 $\pm$ 0.69                                  & \textbf{90.31 $\pm$ 0.17}                                  & \textbf{86.14}                       \\
		\bottomrule
	\end{tabular}
\end{table*}

\subsubsection{Impact of different masking ratios}
The impact of different masking ratios is investigated for the input sequence of the momentum encoder in saliency-guided contrastive learning. In Tabel \ref{table:maskratio}, we tabulate the fine-tuning evaluation performance of SSiT on all three datasets under different masking ratios. In the case of a masking ratio of 0\%, the contrastive loss becomes similar to that in MoCo-v3 without fixing the linear projection layer. As shown in Table \ref{table:maskratio}, with a masking ratio of 25\%, the pre-trained model surpasses models with all other compared masking ratios on both DDR and APTOS2019, whereas the model with a masking ratio of 50\% attains the best result on Messidor-2. A performance drop is observed when the masking ratio is larger than 50\%; a large masking ratio may erase diagnostic patches, leading to incomplete or even misleading target representations for the query encoder. After overall consideration, we finally select 25\% as the default masking ratio in SSiT.

\subsubsection{Generalization ability}
We further characterize the robustness and capability of SSiT to generalize under different ViT variants, including ViT-Ti (ViT Tiny) \cite{touvron2021training} and ViT-B (ViT Base) \cite{dosovitskiy2020image}. ViT-Ti is a lightweight ViT consisting of 12 transformer blocks each with a 3-headed self-attention module and a 192-dimension hidden size for embedding features. ViT-B is a larger model with the same 12 transformer blocks, but each block has a 12-headed self-attention module and a 768-dimension hidden size. Both these two models are trained with the same hyper-parameters as described in Sec. \ref{train_detail}. The consistent improvements across all three evaluation datasets, as tabulated in Table \ref{table:network1}, reveal that SSiT can generalize to different ViT architectures and benefits from a higher capacity in terms of the network architecture. A special case is that for the APTOS2019 dataset under fine-tuning evaluation, no improvement from ViT-B over ViT-S is observed. The reason may be that this dataset is relatively simple and the performance is already saturated.

\subsubsection{Saliency detection methods}
\begin{figure}[t]
	\centering
	\includegraphics[width=0.85\linewidth]{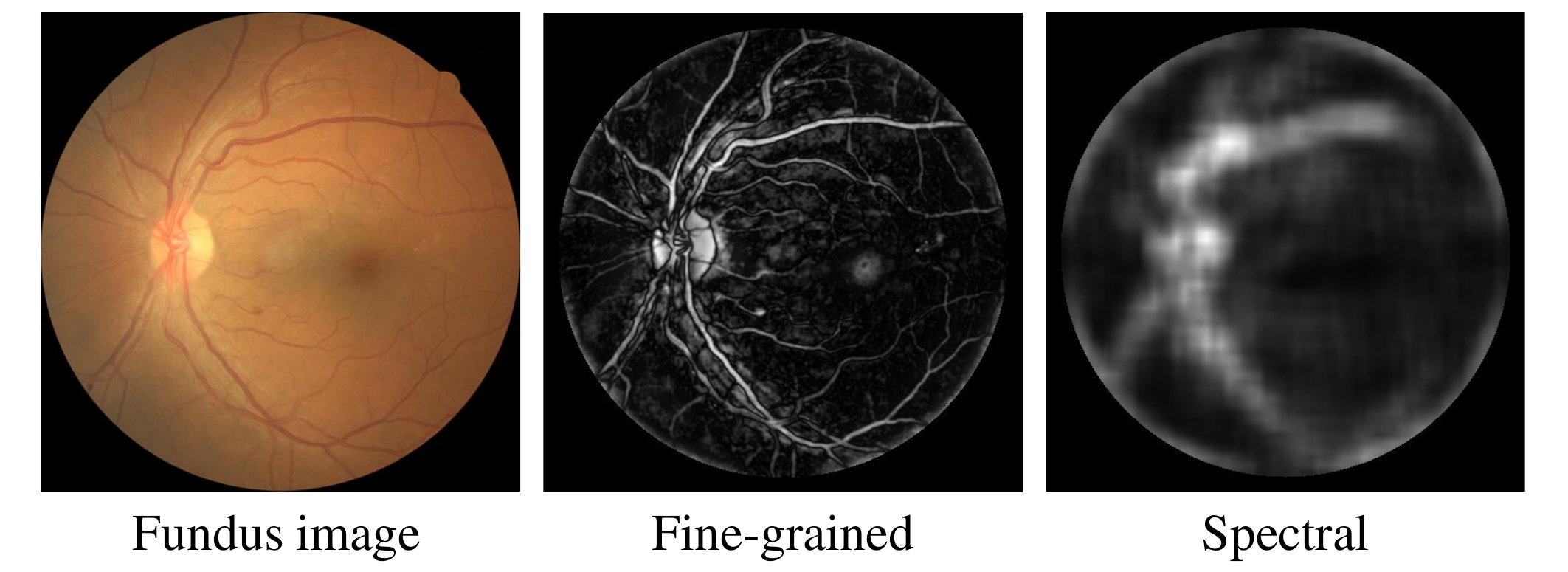}
	\caption{Saliency map comparison from two static detection methods.}
	\label{fig:saliency2}
\end{figure}

A static fine-grained saliency detection method \cite{montabone2010human} is utilized in our proposed SSiT to generate the saliency maps of fundus images without requiring additional training. To analyze the importance of this fine-grained saliency detection method, we compare it with another commonly used static saliency detection method, namely the spectral residual-based saliency detection method \cite{hou2007saliency}. As illustrated in Fig. \ref{fig:saliency2}, the spectral saliency method produces very rough and blurry saliency maps, tending to overlook structural boundary details and disease-related lesions in the fundus images. Table \ref{table:detection} shows that employing saliency map from the spectral method in SSiT results in a decrease in $\kappa$ by 9.97\% on DDR, 10.21\% on Messidor-2 and 1.76\% on APTOS2019, under the fine-tuning evaluation protocol. These comparison results suggest that more accurate saliency maps highlighting regions of interest can greatly enhance the performance of SSiT. It is noteworthy that SSiT even with a coarse saliency map still outperforms other ViT-based SSL methods, demonstrating the efficacy of SSiT in learning representations from saliency maps.

\begin{table}[t]
	\centering
	\caption{The fine-tuning evaluation performance of SSiT with different saliency detection methods. [$\kappa$ (\%)]}
	\label{table:detection}
	\renewcommand{\arraystretch}{1.2}
	\begin{tabular}{lccc}
		\toprule
		Method            & DDR                       & Messidor-2                  & APTOS2019                 \\ \hline
		Spectral saliency              & 71.91 $\pm$ 1.20          & 67.32 $\pm$ 0.57    & 91.21 $\pm$ 0.40         \\
		Fine-grained saliency                & \textbf{81.88 $\pm$ 0.26} & \textbf{77.53 $\pm$ 0.84}            & \textbf{92.97 $\pm$ 0.29} \\
		\bottomrule
	\end{tabular}
\end{table}

\subsubsection{Balance of the two objectives}
We further analyze the impact of the hyper-parameter $\lambda_{seg}$ which controls the relative weight of the saliency segmentation objective in SSiT. In Table \ref{table:lambda}, we tabulate the fine-tuning evaluation performance of SSiT with different $\lambda_{seg}$. In the case of $\lambda_{seg}$ being 0, only the saliency-guided contrastive objective is employed. As shown in that table, the best-performing $\lambda_{seg}$ is 10 (our default). Among all three datasets, we observe pronounced performance enhancement on the Messidor-2 dataset, which is a critically small fundus dataset with only 1748 images. This clearly highlights the importance of learning local fine-grained information in scenarios with limited training data.

\subsection{Visualization}
\subsubsection{Saliency map segmentation}
\begin{figure*}[h]
	\centering
	\includegraphics[width=0.85\linewidth]{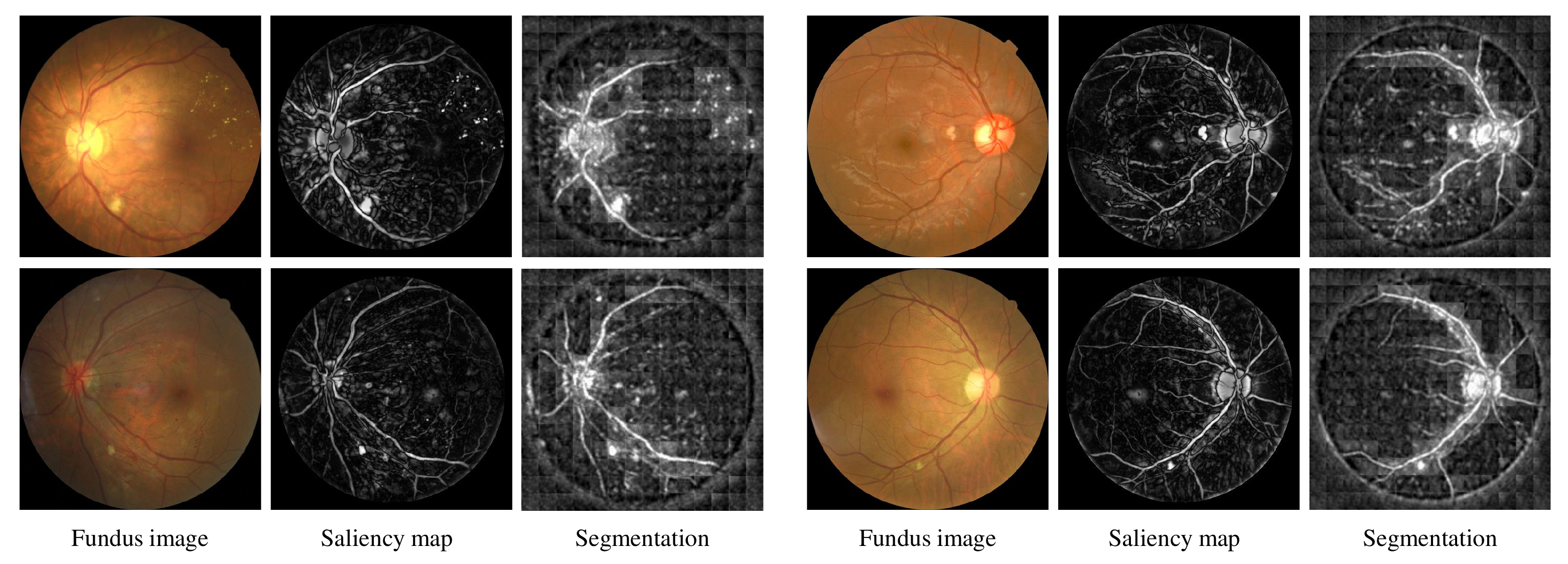}
	\caption{Visualization of saliency map segmentation results on representative DDR samples.}
	\label{fig:segmentation}
\end{figure*}

In Fig. \ref{fig:segmentation}, we visualize the saliency map segmentation results for representative images from the DDR dataset. Apparently, the segmentation predictions clearly highlight retinal structures and DR-related lesions in the fundus images. Notably, our decoder for segmentation is extremely lightweight (only one linear layer). Therefore, the promising segmentation performance implies that the learned representations contain rich fine-grained information, promoting the pre-trained model's detection power for diagnostic features that greatly benefit downstream DR grading.

\begin{table}[t]
	\centering
	\caption{The fine-tuning evaluation performance of SSiT with different $\lambda_{seg}$. [$\kappa$ (\%)]}
	\label{table:lambda}
	\renewcommand{\arraystretch}{1.2}
	\begin{tabular}{cccc}
		\toprule
		$\lambda_{seg}$            & DDR                       & Messidor-2                  & APTOS2019                 \\ \hline
		0                & 74.33 $\pm$ 0.51          & 55.92 $\pm$ 1.49    & 90.32 $\pm$ 0.38         \\
		1                & 78.73 $\pm$ 0.63          & 75.33 $\pm$ 1.01    & 92.86 $\pm$ 0.37         \\
		10               & \textbf{81.88 $\pm$ 0.26} & \textbf{77.53 $\pm$ 0.84}            & \textbf{92.97 $\pm$ 0.29} \\
		20               & 73.83 $\pm$ 0.90          & 66.48 $\pm$ 1.73    & 80.87 $\pm$ 0.84         \\
		\bottomrule
	\end{tabular}
\end{table}

\subsubsection{Self-supervised self-attention map}
\label{attention_map}
\begin{figure*}[tbp]
	\centering
	\includegraphics[width=0.88\linewidth]{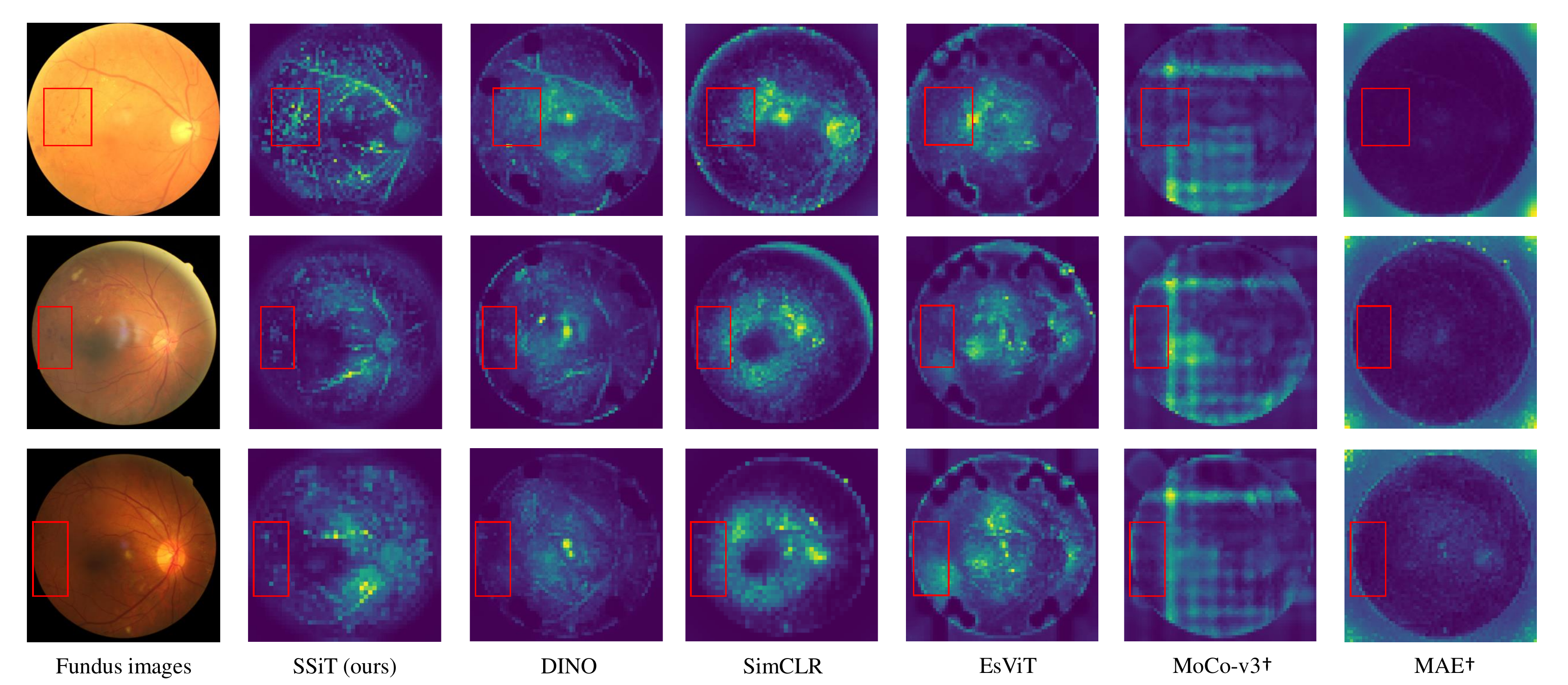}
	\caption{Self-attention maps from different self-supervised ViTs with an input resolution of 1024$\times$1024 and a patch size of 16. $^{\dag}$ MoCo-v3 and MAE fail probably because the input resolution changes sharply.}
	\label{fig:attention}
\end{figure*}

Recently, DINO \cite{caron2021emerging} shows that the attention maps of self-distillation based pre-trained ViTs contain semantic information of natural images, but it is still unclear whether this property applies to medical images and other SSL methods. In Fig. \ref{fig:attention}, we visualize the self-attention maps of different self-supervised ViTs. Fundus images are randomly sampled from the DDR dataset that are not involved in pre-training. Following \cite{caron2021emerging}, in order to generate fine attention maps, images are resized to be of a larger resolution $1024 \times 1024$, resulting in $64 \times 64$ patches with a patch size of 16 as the input sequence. They are fed into self-supervised ViTs and self-attention maps from the last layer of each ViT are depicted. We visualize the attention by taking an average of the normalized multi-headed self-attention maps for the global token $\bm{z}_{class}$. As shown in Fig. \ref{fig:attention}, only the attention maps from SSiT and DINO exhibit scene layouts of the fundus images. However, DINO is likely to overlook DR-related lesions, while SSiT more precisely highlights the corresponding diagnostic regions with clearer boundaries (red boxes in Fig. \ref{fig:attention}). SimCLR and EsViT, on the other hand, focus on bright regions and do not identify any semantic information. MoCo-v3 and MAE fail in displaying meaningful regions. We speculate that since MoCo-v3 fixes the linear projection layer in pre-training and the task in MAE focuses on image reconstruction, their corresponding pre-trained models are very sensitive to increases in the input resolution which induces changes in the input patch distribution. Fig. \ref{fig:attention}'s visualization results demonstrate that with the guidance of saliency map, SSiT successfully incorporates rich saliency information of fundus images, the property of which is not possessed by other contrastive SSL methods. This property of SSiT also contributes to an improvement in the downstream DR grading performance by encouraging the encoder to focus on DR-related diagnostic characteristics.

\begin{table}[t]
	\centering
	\caption{The fine-tuning evaluation performance of SSiT and other ViT-based self-supervised learning methods on Ichallenge-AMD and Ichallenge-PM. [$\kappa$ (\%)]}
	\label{table:external}
	\renewcommand{\arraystretch}{1.2}
	\begin{tabular}{lcc}
		\toprule
		Method            & Ichallenge-AMD                       & Ichallenge-PM                                   \\ \hline
		\textcolor{gray}{Random init.}   & \textcolor{gray}{11.23 $\pm$ 2.60} & \textcolor{gray}{88.07 $\pm$ 1.87} \\
		SimCLR \cite{chen2020simple}       & 63.01 $\pm$ 1.53          & \underline{96.03 $\pm$ 1.12}           \\
		MoCo-v3 \cite{chen2021empirical}               & 66.51 $\pm$ 2.89 & 95.53 $\pm$ 1.86        \\
		DINO \cite{caron2021emerging}             & 58.63 $\pm$ 8.36 & 84.72 $\pm$ 7.27             \\
		MAE \cite{he2022masked}                 & 56.72 $\pm$ 5.38 & 95.52 $\pm$ 5.33             \\
		EsViT \cite{li2021efficient}                                 & \underline{70.46 $\pm$ 3.58} & 94.53 $\pm$ 2.45             \\
		SSiT (ours)                 & \textbf{76.27 $\pm$ 3.20} & \textbf{97.02 $\pm$ 0.10}             \\
		\bottomrule
	\end{tabular}
\end{table}

\subsection{Other Retinal Tasks}
In this section, we show that SSiT can also learn general feature representations for diverse retinal tasks. To analyze the generalizability of learned features, we evaluate our proposed method on two distinct downstream retinal tasks, i.e., age-related macular degeneration (AMD) and pathologic myopia (PM) diagnoses. Typical signs of AMD in fundus images include drusen, exudation and hemorrhage, while those of PM include atrophy and lacquer crack. For evaluation purposes, we employ two publicly available datasets, namely Ichallenge-AMD \cite{amd} and Ichallenge-PM \cite{pm}, which are specifically and respectively associated with AMD and PM. Both datasets contain 400 fundus images with normal and abnormal annotations, each of which is randomly split into 70\%/10\%/20\% for training/validation/testing. As tabulated in Table \ref{table:external}, SSiT consistently outperforms all other ViT-based SSL methods by at least 9.76\% on Ichallenge-AMD and 0.99\% on Ichallenge-PM. Please note that the task of Ichallenge-PM is relatively simple, so there is not much room for improvement. These experimental results suggest that the saliency-guided feature representations learned from fundus images not only benefit DR grading but also facilitate the diagnoses of other retinal diseases. In essence, SSiT effectively learns generalized representations from fundus images, which can be applied to diverse retinal diagnostic tasks.

\subsection{Fundus Segmentation Tasks}
SSiT demonstrates its ability to learn semantic information of DR-related diagnostic regions. To substantiate this capability, we evaluate the fine-tuning performance on three segmentation tasks: vessel segmentation using the DRIVE dataset \cite{staal2004ridge} as well as lesion segmentation and optic disc segmentation using the IDRiD dataset \cite{porwal2018indian}. The official dataset split is employed for training and testing, and the Dice score is utilized to evaluate the segmentation performance.

As depicted in Table \ref{table:segmentation}, SSiT consistently outperforms other ViT-based SSL methods across all segmentation tasks. Notably, on the lesion segmentation task, SSiT outperforms other methods by at least 4.37\% Dice score. This well aligns with the observations in Section \ref{attention_map} that SSiT is the only method highlighting DR-related lesions in self-supervised attention maps. Its superior segmentation performance across all segmentation tasks demonstrates SSiT's capability of learning fine-grained features that can be effectively transferred to dense downstream tasks.

\begin{table}[t]
	\centering
	\caption{The fine-tuning segmentation performance of SSiT and other ViT-based self-supervised learning methods. [Dice (\%)]}
	\label{table:segmentation}
	\renewcommand{\arraystretch}{1.2}
	\begin{tabular}{lccc}
		\toprule
		Method            & DRIVE-Vessel                       & IDRiD-Lesion                       & IDRiD-OD                                   \\ \hline
		\textcolor{gray}{Random init.}   & \textcolor{gray}{74.84 $\pm$ 0.05} & \textcolor{gray}{15.91 $\pm$ 3.85} & \textcolor{gray}{88.61 $\pm$ 2.04} \\
		SimCLR \cite{chen2020simple}        & 75.23 $\pm$ 0.11          & 31.23 $\pm$ 3.36          & 92.95 $\pm$ 0.53           \\
		MoCo-v3 \cite{chen2021empirical}               & 75.34 $\pm$ 0.10 & 43.70 $\pm$ 1.61 & 93.55 $\pm$ 0.48        \\
		DINO \cite{caron2021emerging}             & 75.82 $\pm$ 0.01 & 49.49 $\pm$ 1.00 & \underline{94.76 $\pm$ 0.37}             \\
		MAE \cite{he2022masked}                 & \underline{77.19 $\pm$ 0.07} & \underline{54.64 $\pm$ 1.65} & 94.18 $\pm$ 0.38             \\
		EsViT \cite{li2021efficient}                                  & 75.13 $\pm$ 0.11 & 39.78 $\pm$ 1.20 & 92.10 $\pm$ 0.32             \\
		SSiT (ours)                 & \textbf{78.05 $\pm$ 0.01} & \textbf{59.01 $\pm$ 2.75} & \textbf{95.32 $\pm$ 0.36}             \\
		\bottomrule
	\end{tabular}
\end{table}

\begin{table*}[t]
	\centering
	\caption{The linear performance of models pre-trained with different objectives of SSiT. SCL and Seg respectively denote the saliency guided contrastive loss and the saliency map segmentation loss. The best ones are bolded while the second best ones are underlined.}
	\label{table:role}
	\renewcommand{\arraystretch}{1.2}
	\begin{tabular}{cccccccc}
		\toprule
		\multicolumn{2}{c}{Components} & \multicolumn{3}{c}{DR grading [Kappa (\%)]}                                                & \multicolumn{3}{c}{Segmentation [Dice (\%)]}                                                  \\
		\cmidrule(lr){1-2} \cmidrule(lr){3-5} \cmidrule(lr){6-8} 
		SCL            & Seg           & DDR                       & Messidor2                 & APTOS2019                 & DRIVE-Vessel              & IDRiD-Lesion              & IDRiD-OD                  \\ \hline
		\ding{55}      & \ding{55}     & 15.56 $\pm$ 5.60          & 15.80 $\pm$ 7.14          & 57.99 $\pm$ 2.19          & 19.50 $\pm$ 0.91          & 9.55 $\pm$ 0.60           & 83.04 $\pm$ 0.40          \\
		\ding{51}      & \ding{55}     & \underline{66.56 $\pm$ 0.51}         & \underline{58.12 $\pm$ 0.62}          & \underline{87.28 $\pm$ 0.11}          & 21.32 $\pm$ 0.15          & 23.42 $\pm$ 0.60          & 87.72 $\pm$ 0.02          \\
		\ding{55}      & \ding{51}     & 62.45 $\pm$ 1.28          & 36.13 $\pm$ 1.94          & 81.09 $\pm$ 0.27          & \underline{57.23 $\pm$ 0.06}          & \underline{38.36 $\pm$ 0.15}          & \underline{91.22 $\pm$ 0.06}          \\
		\ding{51}      & \ding{51}     & \textbf{71.33 $\pm$ 0.78} & \textbf{67.23 $\pm$ 0.53} & \textbf{89.65 $\pm$ 0.20} & \textbf{78.05 $\pm$ 0.01} & \textbf{59.01 $\pm$ 2.75} & \textbf{95.32 $\pm$ 0.36} \\ \bottomrule
	\end{tabular}
\end{table*}

\subsection{Role of Each Objective}
\label{role}
There are two learning objectives in SSiT, namely saliency-guided contrastive learning and saliency map segmentation. The first objective aims to maximize the similarity between the global representation of the query encoder and that of the saliency-guided key encoder. This task mainly focuses on learning discriminative representations at the global level. In contrast, the second objective requires the model to predict salient regions at the pixel level, thereby enhancing fine-grained and local semantics.

To further analyze the role of each objective in SSiT, we evaluate the corresponding models pre-trained with different self-supervised objectives on both downstream classification and segmentation tasks under the linear evaluation setting. Linear downstream segmentation performance directly reflects the pre-training quality in extracting local features. As depicted in Table \ref{table:role}, the model pre-trained solely with saliency-guided contrastive learning significantly outperforms that pre-trained solely with saliency map segmentation on downstream classification tasks. Conversely, the reverse trend is observed on downstream segmentation tasks. These results suggest that saliency map segmentation contributes to the enhancement of local feature representations, encouraging the pre-trained model to focus on fine-grained semantics. On the other hand, saliency-guided contrastive learning improves the pre-trained model in extracting global discriminative representations. By combining these two objectives, SSiT allows the pre-trained model to simultaneously extract global discriminative and local fine-grained representations from fundus images, and thus outperforms other SSL methods on all evaluation tasks.

\section{Discussion}
\label{section5}
Motivated by its huge success in computer vision, SSL recently has been actively explored in medical image analysis. However, unlike natural images in computer vision, medical images are much more expensive to collect, more complicated, and more informative. Therefore, the general SSL approaches in computer vision may be incompatible or perform unsatisfactorily for medical image analysis. Here, we emphasize the importance and effectiveness of incorporating prior knowledge into SSL for medical images, such as the saliency information that provides diagnostic characteristics for downstream DR grading in this work. Introducing prior knowledge into SSL may enhance disease-related features in the learned representations and alleviate the demanding requirement of a large number of samples in the pre-training step, thereby improving the feasibility and generalizability of self-supervised learning in the medical image analysis realm. Note that with an extra pre-training dataset, as tabulated in Table \ref{table:ddr}, our proposed SSiT achieves SOTA DR grading performance on the DDR dataset which has an official test split.

Although we have only validated the effectiveness of SSiT in learning representations from fundus images, our method can be easily transferred to other types of medical images since fine-grained information is very important and is held in almost all types of medical imaging data, which will be one of our future extensions. As described in Sec. \ref{gsm}, the saliency detection method can be applied to other types of medical images such as OCT, MRI and chest X-ray which also have clear intensity differences between objects of interest and background. With that being said, SSiT may have limited effectiveness for images of complex backgrounds such as whole slide images, because the saliency may be difficult to be identified in those images. Another limitation of our proposed SSiT framework, which is also common to other SSL methods for medical images, is that a large amount of unlabeled data is still needed to train the self-supervised model (\eg 88,702 fundus images in this work and \cite{li2020self}, 112,120 x-ray images in \cite{haghighi2022dira} and 454,295 dermatology images in \cite{azizi2021big}). The requirement for large-scale datasets may be difficult to meet for some specific types of medical images or modalities, limiting the practical value and breadth of SSiT. We believe that equipping SSiT with more domain-specific prior knowledge may relieve the requirement of large-scale pre-training image datasets. Therefore, one of our future research directions is to explore more effective methods of incorporating domain knowledge into SSiT, and thus further reducing the sample size of the pre-training dataset employed in SSiT.

\begin{table}[t]
	\centering
	\caption{Comparison results with SOTA DR grading methods on the DDR dataset. '$\dagger$' and '$\ddagger$' respectively denote the results are reported in \cite{cabnet} and \cite{wang2021joint}.}
	\label{table:ddr}
	\renewcommand{\arraystretch}{1.2}
	\begin{tabular}{llc}
		\toprule
		Method      & Backbone                        & $\kappa$ (\%)                                   \\ \hline
		DenseNet-121$\textsuperscript{\textdagger}$ \cite{huang2017densely}     & DenseNet-121                 & 74.4             \\
		CABNet \cite{cabnet}      & DenseNet-121             & 78.6             \\
		AFN$\textsuperscript{\textdaggerdbl}$ \cite{lin2018framework}       & Customized arch.    & 74.9 \\
		DeepMT-DR \cite{wang2021joint}  & Customized arch.   & 80.2  \\
		SSiT (ours)   & ViT-S    & 81.9  \\
		SSiT (ours)   & ViT-B    & \textbf{83.8}  \\
		\bottomrule
	\end{tabular}
\end{table}

\section{Conclusion}
\label{section6}
In this work, we present and validate a novel self-supervised learning framework, namely SSiT, to learn generalizable and transferable representations from fundus images. SSiT distinguishes itself from other SSL methods by introducing saliency maps into the self-supervised paradigm. In the proposed contrastive learning scheme, trivial patches are removed from the input sequence of the momentum encoder according to the saliency map and thus constrain the momentum encoder to provide target representations focusing on salient regions. As such, the query encoder is guided to focus on DR-related diagnostic regions in fundus images. Furthermore, the query encoder is trained to predict saliency segmentation of the fundus images, which encourages the preservation of fine-grained information in the image representations. Extensive experiments on multiple fundus image datasets are conducted to evaluate the quality of the learned representations, including fine-tuning evaluation, linear evaluation and $k$-NN classification. Experimental results show that SSiT consistently outperforms other SSL methods for DR grading. We also show that the self-supervised ViTs in SSiT deliver rich semantic information of DR-related diagnostic characteristics, which is not observed in other SSL methods.

\section*{References}
\balance
\bibliographystyle{ieeetr}
\bibliography{./ref.bib}

\end{document}